\documentclass[letterpaper, 10 pt, journal, twoside]{IEEEtran}
\bstctlcite{bstctl:etal, bstctl:nodash, bstctl:simpurl}
\usepackage{xcolor}
\usepackage{times}
\usepackage{graphicx}
\usepackage{amsmath}
\usepackage{amssymb}
\usepackage{textcomp}
\usepackage{multirow}
\usepackage{booktabs}
\usepackage{flushend}
\usepackage{xcolor}
\definecolor{black}{rgb}{0.0,0.5,0.0}
\usepackage{multicol}
\usepackage{color}
\usepackage{makecell}
\usepackage{bbding}
\usepackage{array}
\usepackage{pifont}
\usepackage{colortbl}

\usepackage{epsfig}
\usepackage{adjustbox} % in the preamble
\usepackage{textcomp}
\usepackage{adjustbox}

\usepackage{hhline}
\usepackage{subcaption}
\usepackage{enumitem}
\usepackage{verbatim}
\usepackage{cite}

\usepackage{tabularx} % Required for tabularx environment
\usepackage{enumitem} % Required for customizing itemize environment (optional)
\newcolumntype{b}{X}
\newcolumntype{s}{>{\hsize=.1\hsize}X}

\usepackage{gensymb}

\usepackage{wrapfig}

\usepackage[figureposition=bottom,tableposition=top]{caption}
\usepackage[colorlinks,pagebackref=false,citecolor=blue,bookmarks=false,hypertexnames=true]{hyperref}

\ifx00 %ifx01 to turn off and ifx00 to turn on
\newcommand{\sy}[1]{\textcolor{red}{[SY: #1]}}
\newcommand{\AD}[1]{\textcolor{red}{[Arindam: #1]}}
\newcommand{\brk}[1]{\textcolor{red}{[BRK: #1]}}
\else 
\newcommand{\sy}[1]{\textcolor{red}{}}
\newcommand{\AD}[1]{\textcolor{red}{}}
\newcommand{\brk}[1]{\textcolor{red}{}}
\fi

\bstctlcite{IEEEexample:BSTcontrol}

\begin{document}

% \title{
% Guidelines for Creating a Safe Dataset: A Framework for AI-Based Perception in Autonomous Driving
% }

% \title{
% Safe AI Dataset for Autonomous Driving: \\Overview, Guidelines and Challenges
% }

\title{
Dataset Safety in Autonomous Driving: Requirements, Risks, and Assurance
}

\author{Alireza Abbaspour$^{1}$, Tejaskumar Balgonda Patil$^{1}$, B Ravi Kiran$^{1}$, Russel Mohr$^{1}$, Senthil Yogamani$^{1}$\\
$^1$Qualcomm Inc
}

\markboth{IEEE Transactions on Intelligent Transportation Systems}
{Abbaspour \MakeLowercase{\textit{et al.}}: Dataset Safety in Autonomous Driving: Requirements, Risks, and Assurance
}

\maketitle

\begin{abstract}
Dataset integrity is fundamental to the safety and reliability of AI systems, especially in autonomous driving. This paper presents a structured framework for developing safe datasets aligned with ISO/PAS 8800 guidelines. Using AI-based perception systems as the primary use case, it introduces the AI Data Flywheel and the dataset lifecycle, covering data collection, annotation, curation, and maintenance. The framework incorporates rigorous safety analyses to identify hazards and mitigate risks caused by dataset insufficiencies. It also defines processes for establishing dataset safety requirements and proposes verification and validation strategies to ensure compliance with safety standards. In addition to outlining best practices, the paper reviews recent research and emerging trends in dataset safety and autonomous vehicle development, providing insights into current challenges and future directions. By integrating these perspectives, the paper aims to advance robust, safety-assured AI systems for autonomous driving applications.
\end{abstract}

\IEEEpeerreviewmaketitle

\section{Introduction}

The rapid advancement of autonomous driving technology has brought about significant transformations in the transportation sector. Autonomous vehicles (AVs) promise to enhance road safety, improve traffic efficiency, and provide new mobility solutions \cite{mirzarazi2024safety}. It makes use of a diverse set of AI tasks and custom applications to make it robust and safe for the consumer \cite{sistu2019neurall, kumar2018near, yahiaoui2019overview, chennupati2019auxnet}. However, the deployment of these systems depends critically on the integrity and reliability of the datasets used to train and validate AI models. Deficiencies in these datasets can lead to catastrophic failures in real-world scenarios, making dataset safety a central concern \cite{wang2025collaborative}. 

 Recent advancements in late 2024 and throughout 2025 have shifted the focus toward a "data-centric" AI paradigm, where the quality and representativeness of data are viewed as safety-critical artifacts equivalent to traditional software code \cite{su2025robosense, truckscenes2024, ni2025lane, wang2025uniocc}. This shift is underscored by the emergence of the ISO/PAS 8800 standard, which provides a roadmap for managing the unique risks of AI, such as performance degradation in scenarios not represented in training distributions. Furthermore, industry leaders have identified "Physical AI" as the next frontier, requiring datasets that go beyond perception to encompass reasoning, planning, and real-time interaction with the physical world.

Reliable autonomous driving systems depend critically on the quality and diversity of training datasets. These extensive data resources help ensure that Advanced Driving Assistance Systems (ADAS) components remain robust in complex, real-world environments \cite{liu2024survey}. The rise of autonomous driving datasets has paralleled the development of self-driving technologies to the point that datasets are now considered one of the core building blocks on the path toward full autonomy. Preparing datasets through collecting, cleaning, annotating, and augmenting data has become a cornerstone of ADAS development, directly impacting the performance and safety of learned driving functions \cite{janai2020computer,gao2021we}. 

The period between 2024 and 2026 has seen the release of specialized datasets targeting overlooked domains, such as low-speed unmanned vehicles and heavy autonomous trucks, which face unique perception challenges in near-field and long-range environments. For instance, the introduction of 4D radar data in datasets like MAN TruckScenes \cite{truckscenes2024} has redefined the standards for long-range environmental awareness, providing 360-degree coverage essential for heavy-duty vehicle safety.

Building on this foundation, recent efforts have focused on ensuring dataset diversity and annotation quality. The surge in published autonomous driving datasets and dedicated surveys between 2020 and 2024 reflects a growing recognition that diverse and representative data are essential for training ADAS perception and decision modules \cite{liu2024survey, joseph2021autonomous}. Diversity in sensor modalities (camera, LiDAR, radar, etc.) and environmental conditions (lighting, weather, road types) enhances the generalizability of ADAS algorithms across scenarios \cite{borse2023x, klingner2023x3kd, rashedfisheyeyolo}. However, the emerging trend of early fusion systems introduces additional complexity in the design and assurance of multimodal datasets.  In response, the research community has developed "unified benchmarks" like UniOcc \cite{wang2025uniocc}, which standardize multi-modal data across multiple real-world and synthetic datasets to enable cross-domain evaluation and better handle out-of-distribution generalization. These unified frameworks allow researchers to analyze the "semantic gap" that exists between different sensor configurations and operational domains \cite{ullrich2026toward, zhou2026novel}.

 Despite the growing literature on autonomous driving datasets, most existing works primarily focus on dataset scale, modality coverage, or task-specific benchmarking, while providing limited discussion on systematic dataset safety assurance and lifecycle management. In particular, the operationalization of dataset safety properties into measurable constraints, traceable requirements, and verifiable evidence remains insufficiently explored in the literature. This gap motivates the need for structured approaches that connect dataset engineering practices with safety analysis methodologies and emerging automotive safety standards.

Well-prepared datasets that cover a wide range of driving situations help prevent models from becoming brittle or biased toward narrow distributions. For example, Gao et al. survey the landscape of 3D LiDAR segmentation datasets and question whether current data suffices for robust semantic perception, highlighting the need for more comprehensive coverage of driving scenes \cite{gao2021we}. Similarly, ensuring consistency and accuracy in data labeling has been identified as a key factor in dataset quality—recent studies emphasize the importance of standardized annotation pipelines to maintain uniform labels across millions of frames \cite{liu2024survey}. Without careful curation and annotation, even large datasets may fail to improve ADAS reliability due to systematic errors or omissions \cite{uricarvisapp19}. 

 Recent breakthroughs in automated labeling platforms like Encord \cite{clough2025avlabelingplatforms} have integrated Foundation Models such as SAM2 to facilitate precise segmentation across 2D, 3D, and video data, significantly reducing manual effort while improving label consistency. Moreover, the adoption of "human-in-the-loop" workflows ensures that expert verification remains a critical safety check for complex scenarios like multi-agent occlusions and adverse weather conditions.
Concrete examples further illustrate the importance of robust datasets in autonomous driving. The Waymo Open Dataset, for instance, has been instrumental in advancing research by providing high-quality data for perception and planning tasks \cite{hamon2022artificial}. Similarly, the Autonomous Vehicles: Timeline and Roadmap Ahead report highlights the need for comprehensive safety analyses and rigorous validation processes to mitigate risks associated with dataset insufficiencies \cite{mirzarazi2024safety}.

 To bridge the gap between static datasets and real-world reasoning, the DriveLMM-o1 dataset, released in 2025 \cite{ishaq2025drivelmm}, provides explicit step-by-step reasoning annotations, enabling models to "think" through perception, prediction, and planning tasks rather than relying solely on pattern recognition. This shift toward reasoning-centric datasets is critical for ensuring compliance with complex traffic rules and improving safety in long-tail scenarios. To address these concerns, researchers have proposed guidelines for improving AI safety through safe dataset preparation. Mirzarazi et al. \cite{mirzarazi2024safety} discuss the potential safety risks of deploying deep neural network classifiers in ADAS and offer recommendations for mitigating these risks. Wang et al. \cite{wang2025collaborative} focus on collaborative perception datasets, emphasizing the importance of multi-agent information fusion to enhance perception accuracy and safety. Additionally, the Joint Research Center's analysis \cite{hamon2022artificial} provides insights into the vulnerabilities of AI components in automated driving and suggests strategies to reduce these risks. 

 By 2026, generative world models such as Sora 2 \cite{zheng2025open} and NVIDIA Cosmos \cite{agarwal2025cosmos} have emerged as pivotal tools for "scenario dreaming," where vehicles rehearse potential future trajectories in a compact latent state to detect hazards before they manifest in reality. These world models provide high-fidelity representations of temporal dynamics and causal contexts, effectively serving as learned simulators for safe planning and decision-making. Recognizing these challenges, the research community has devoted significant effort to cataloging and analyzing ADAS datasets. Janai et al. (2020) provided an early comprehensive overview of computer vision problems in autonomous vehicles, emphasizing the role of datasets in advancing the state of the art \cite{janai2020computer}. In subsequent years, numerous surveys and reviews have illuminated different facets of dataset preparation. Liu et al. (2021) examined dozens of driving datasets, analyzing their sensor modalities, scales, and supported tasks \cite{liu2021survey}. Their study also explored techniques for bridging the gap between simulated and real-world data, including domain adaptation strategies and automatic labeling methods to efficiently expand training sets. As the field evolved, Li et al. (2023) highlighted the emerging open-source data ecosystem for autonomous driving, showcasing the community’s push toward collaborative data sharing and continuously updated dataset platforms \cite{li2023open}. 

More recent literature has introduced task-centric frameworks, such as the Vase Framework, which explicitly map data quality (DQ) metrics to specific task requirements and performance goals \cite{zhou2026novel}. This approach allows for a structured evaluation of how sensor redundancies and environmental factors impact the trustworthiness of real-time decision-making systems. In addition to general surveys, specialized literature has focused on particular aspects of dataset preparation that are critical for ADAS. Synthetic data generation has gained prominence as a means to supplement real-world driving data. Synthetic datasets can produce rare or hazardous scenarios at scale, filling gaps that real data may not cover and reducing the cost of data collection \cite{song2023synthetic}. Bogdoll et al. (2023) review perception datasets tailored for anomaly detection in autonomous driving, noting that capturing out-of-distribution events requires dedicated data collection efforts \cite{bogdoll2023perception}.

 The Adver-City dataset, for example, utilizes the CARLA simulator to recreate dangerous road configurations based on real accident reports, introducing conditions like intense glare and heavy fog to challenge the limits of collaborative perception models \cite{karvat2025adver}. Moreover, dataset preparation plays a vital role beyond perception, extending to higher-level driving intelligence. Decision-making and planning modules in advanced automation also rely on data-driven learning. Wang et al. (2023) show that the performance of decision-making algorithms is strongly influenced by the datasets used for training and validation \cite{wang2024survey}. 

 The emergence of the Vision-Language-Action (VLA) paradigm in 2025 and 2026 seeks to unify these modules, using multi-modal foundation models to transform visual inputs directly into driving trajectories while providing natural language explanations for the decisions made \cite{hu2025vision}. While these works provide valuable insights into dataset characteristics and development practices, limited efforts have systematically integrated dataset lifecycle management, safety analysis techniques, and verification strategies into a unified assurance framework. Furthermore, existing surveys often lack concrete demonstrations of requirement traceability from safety goals to dataset-level metrics, which is critical for safety argumentation and residual risk management.

Taken together, literature from 2020 to 2024 underscores that meticulous dataset preparation is central to ADAS development. The integration of automated MLOps pipelines and regulatory compliance tools has become essential to handle the massive volumes of multisource data required for next-generation L4 and L5 systems.  To address these research gaps, this paper proposes a structured dataset assurance framework that integrates dataset lifecycle concepts, safety analysis methods, and verification strategies into a unified data-centric safety perspective. Unlike existing standards that provide high-level guidance, the proposed approach operationalises dataset safety properties into measurable dataset requirements and key performance indicators (KPIs), enabling traceability and systematic verification.

This paper aims to present a comprehensive framework for creating safe datasets aligned with ISO/PAS 8800 guidelines. Using an End-to-End (E2E) AI-based ADAS as the primary use case, we introduce a structured approach to the data engine and the dataset safety lifecycle, encompassing data collection, annotation, curation, and maintenance. The main contributions of this paper are summarised as follows:\begin{itemize}\item A taxonomy and critical synthesis of dataset lifecycle and assurance approaches in autonomous driving.\item A structured dataset safety framework that operationalises ISO/PAS 8800 dataset properties into measurable constraints.\item A traceability model linking AI safety requirements to dataset KPIs and verification evidence.\item A worked case study demonstrating the application of safety analysis methods to derive actionable dataset requirements.\end{itemize}

The remainder of this paper is organized as follows: Section II introduces the concept of the data flywheel for E2E autonomous driving and reviews recent developments in the field. Section III presents the dataset safety lifecycle, detailing each phase from data acquisition to long-term maintenance. Section IV defines dataset safety requirements, establishing criteria for safe and reliable data use. Section V discusses dataset design principles. Section VI focuses on dataset implementation. Section VII reviews safety analysis methods. Section VIII outlines verification and validation methods. Finally, Section IX concludes the paper and discusses future directions.

\section{Data Flywheel for E2E-AD}
%-------------------------------------------------------------------------
\begin{figure*}
\centering
    \includegraphics[width=0.75\linewidth]{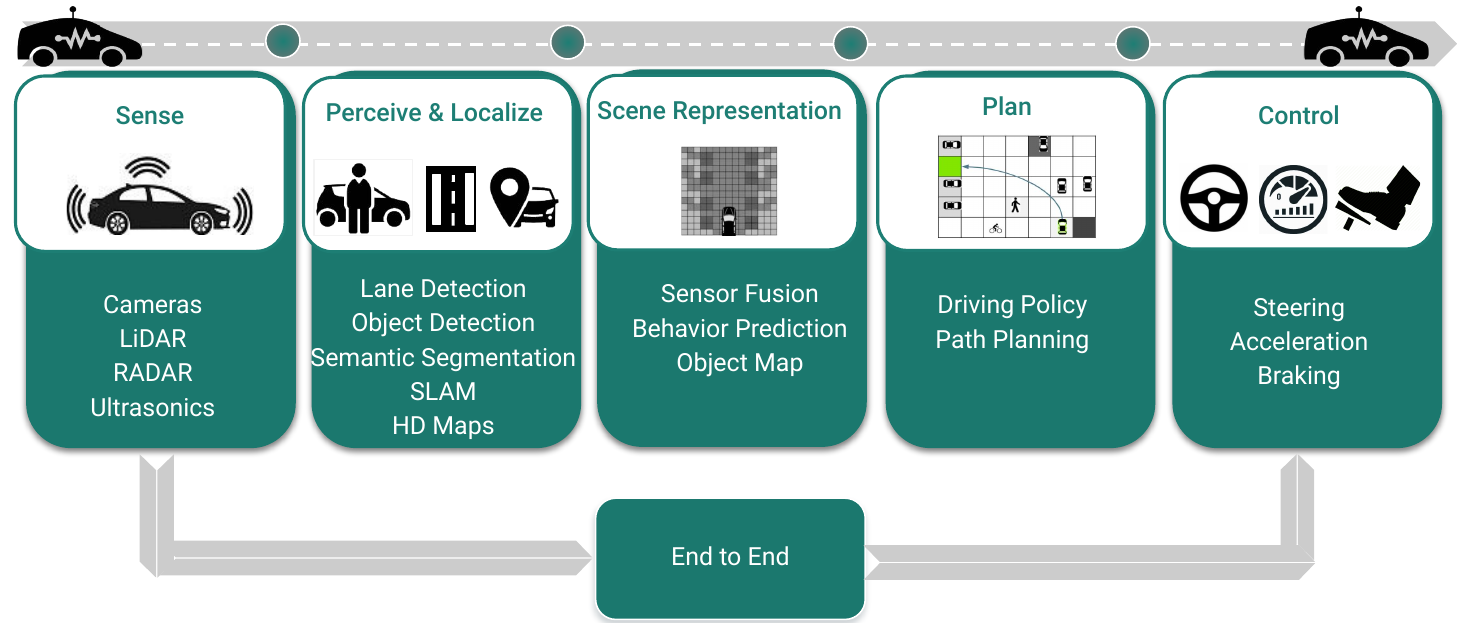}
   \caption{Components of a typical Autonomous Driving Pipeline.}
\label{fig:ad_pipeline}
\end{figure*}
%-------------------------------------------------------------------------
The modern data-pipeline today follows the concept of a data engine or data flywheel (example for Object detection \cite{liang2024aide}) that enables the creation, analysis and curation  of large scale datasets for perception and E2E driving. The data flywheel refers to a machine learning pipeline that enables the identification of mispredictions or labelling errors in a production environment. These flagged instances are sent back to the annotation system for relabelling, after which the model is retrained with the updated data and redeployed. This continuous loop leads to a progressively improving model and dataset.  By early 2026, this paradigm has reached unprecedented scales; for instance, Tesla's data engine has ingested over 9 billion miles of supervised driving data, targeting a 10-billion-mile threshold to achieve safe unsupervised  \cite{TeslaFSD_Safety}. Similarly, Waymo's "outer learning loop" utilizes a sharp Critic model to automatically flag suboptimal driving behaviors from over 100 million autonomous miles, feeding these edge cases back into a closed-loop simulator for reinforcement learning \cite{waymo_demonstrably_safe_ai_2025}.

\begin{figure*}
    \centering
    \includegraphics[width=0.95\linewidth]{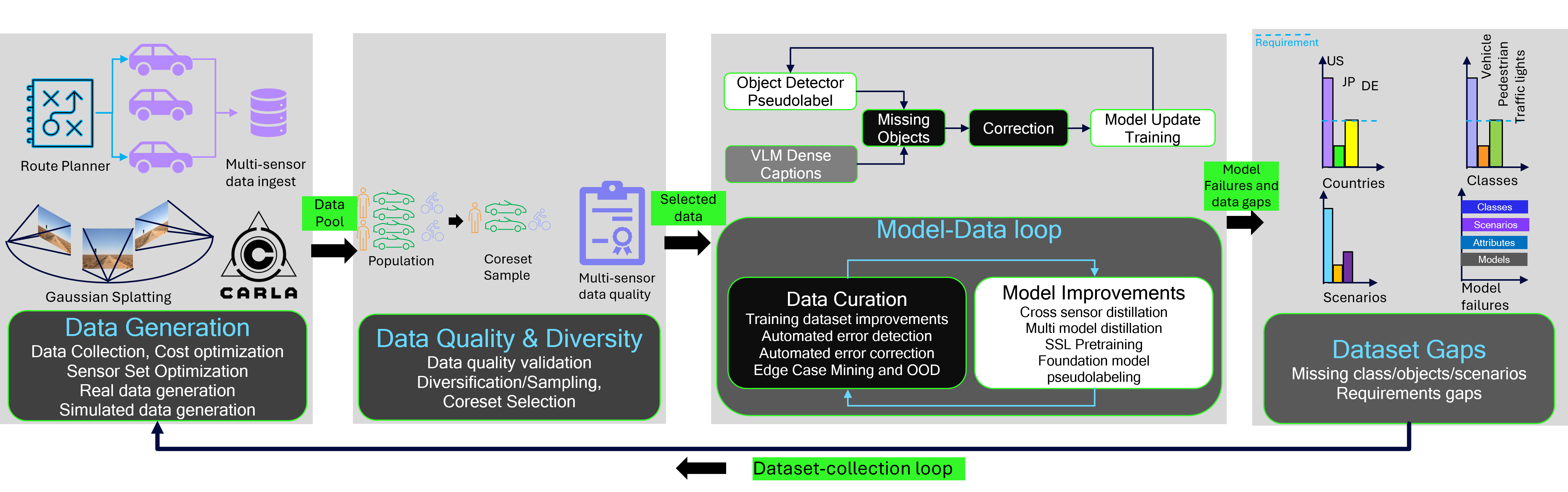}
    \caption{Data flywheel from collection, Data quality and diversification, model training, automated labeling model based annotation (camera+LiDAR DNN), to automated annotation quality check. }
    \label{fig:data-annotation-pipeline}
\end{figure*}

This section describes the steps in a typical data engine components of an autonomous driving system. We have represented our typical data flywheel in figure \ref{fig:data-annotation-pipeline}. Data flywheel or data engines are the manifestation of data centric development where dataset curation has been prioritized over model focussed development \cite{li2024data}. Data engines have developed in the past 5 years across the AD and dataset creation industry Tesla data engine \cite{Tesla_dataengine2022}, Scale AI \cite{scale_ai_dataengine2023}, Aurora \cite{Aurora_dataengine2021}, Momenta Data driven Flywheel \cite{momenta_dataengine2023CVPR}.  Recent additions include the Rivian Data Flywheel, which leverages real-world 3D LiDAR point clouds from customer fleets to power end-to-end Physical AI \cite{RivianUberRobotaxi2026}, and NVIDIA's Alpamayo \cite{wang2025alpamayo}, an open-source "teacher" model suite designed to accelerate the development of reasoning-based autonomous systems.

The key components in building a data pipeline for an autonomous driving systems, perception to planning constitute of the following stages: Dataset requirements and specifications, sensor set choice and configuration analysis, dataset collection and route planning, automated data selection using VLMs, multimodal perception dataset annotation pipelines, automated annotation quality assurance, dataset management and tracking. The offboard annotation pipelines are auto-labeling models trained to output 3D object detection, 3D lane detection, Traffic light, Traffic Sign, Freespace detection. While modern AD systems have moved to performing E2E automated driving described below.

\subsection{E2E driving architectures and datapipelines}
E2E automated driving has progressed to become an achievable engineering target with several subsystems reaching high maturity, e.g. 3D perception models for object detection, online map prediction \cite{liao2022maptr}, motion and path planning with imitation learning. The field is now transitioning toward "Generalist Systems" that unify these modules into a single differentiable framework, reducing information loss and cumulative errors inherent in sequential manual ordering.

There are different families of E2E driving systems. We categorize them as the following:
\begin{enumerate}
    \item Modular E2E driving systems with non differentiable input/output interfaces. Example UniAD \cite{hu2023planning}, Vectorized representations for Autonomous driving VAD \cite{jiang2023vad}. There are also systems that have parallelized or have jointly predicted perception, prediction, mapping and planning  e.g. PARA-Drive \cite{weng2024drive}. A notable 2026 advancement is DriveMamba, which replaces traditional Transformer decoders with a Mamba-based State Space Model (SSM) to achieve linear complexity and better scalability for long-term temporal fusion \cite{su2026drivemamba}.
    \item E2E driving systems without explicit perception tasks e.g. \cite{waywe2024lingo} and \cite{linavigation}. The key goal in such systems are to achieve robust planning without complete perception outputs.
    \item Vision Language Models (VLM) based autonomous driving systems that build upon the usage of VLMs \cite{bordes2024introduction} are today capable of performing perception and motion planning tasks \cite{tian2024drivevlm}. Following trends with robotics, AD systems use VLAs (Vision language action) to predict trajectories condition on vision language features \cite{arai2025covla}.  Emerging VLA models like AutoVLA \cite{zhou2025autovla} and Alpamayo-1 \cite{wang2025alpamayo} employ a "System 1 and System 2" architecture—balancing "fast thinking" for immediate reactive control with "slow thinking" for semantic, chain-of-thought reasoning in rare or complex scenarios.
\end{enumerate}

\subsection{Sensor set selection and configuration analysis}

In the sensing stage, the vehicle collects information about the surroundings via sensors like 
Camera, LiDAR, RADAR, and Ultrasonics. Perception involves the extraction of useful information from 
the raw data and its the most studied module in the literature. Sensor set selection and placement determines the quality of perception system, blind spots, range and precision of 3D detection and 3D lanes. Optimization of placement of sensors is a non trivial problem \cite{kiraz2024survey} involving trade-off between performance, computational resources and cost.

In this section we review the different sensors exist in a modern E2E AV driving system. A brief summary is provided in the Table \ref{tab:sensors}. In the table for each sensor we also compile key issues and safety impact.

\begin{table*}[h!] % [h!] tries to place the table "here"
    \centering
    \caption{Sensors in autonomous driving system}
    \label{tab:sensors}
    \begin{tabularx}{\textwidth}{|s|b|} % X columns automatically adjust width
        \hline
        \textbf{Sensor} & \textbf{Description} \\
        \hline
        Ultrasonic
        &
        \begin{itemize}[nosep, leftmargin=*]
            \item \textbf{Positions}: 6 front, 6 back Door opening, underbody, Range 7cm, 4-5m, increase to 10m for next gen, opening angle 75 degrees
            \item \textbf{Use-case}: Dominant sensor for parking and near field warning 
            \item \textbf{Technology}: Piezoelectric effect to convert vibration to sound, very cheap, pulse BPSK modulation on baseband about sound freq, 20 ms sampling  
            LIN interface, complex envelope signal tof, shape of signal, area, etc, trilateration of multi-sensors to get depth  
            \item \textbf{Impact}: Sound absorbent materials cause issues like foam, clothes. Height classification is coarse, Has issues with sloped roads.
        \end{itemize}
        \\
        \hline
        RADAR
        &
        \begin{itemize}[nosep, leftmargin=*]
            \item \textbf{Positions}: front, rear , 4 corners, \textbf{Range}: 70-100m, 200m, \textbf{FOV} 130 hfov, polar coordinates with azimuth and elevation angle  
        \item \textbf{Usecase}: ACC, CTA, lane change, blind spot, could be used for near range 
        \item \textbf{Technology}: Range and Doppler for velocity of objects, range-doppler vs power spectral graphs are used, penetrates through cars 
        \end{itemize}
        \\
        \hline
        LIDAR
        &
        \begin{itemize}[nosep, leftmargin=*]
            \item \textbf{Position} Center on top of vehicle, front of vehicle, sides/blind spots of vehicle,high range 100-250m, hfov - 130, works in night/day
            \item \textbf{Technology}: Precise range measurement with TOF, Above or below light frequency radiation
            \item \textbf{Types}: 1) Flash lidar - near field - Continental 2) Scanning lidar - Ouster/Velodyne 128/64/32 beams, Scala 4/8 beams
            \item \textbf{Issues}: Affect eyes, multiple lidar interference, poor performance in bad weather: fog, heavy rain, heavy snow
            \item \textbf{Perception limitations}: sparsity issues causes small objects not detected properly
        \end{itemize}
        \\
        \hline
        Cameras
        &
        \begin{itemize}[nosep, leftmargin=*]
            \item \textbf{Sensor configurations}: V-Cameras, 1V, 5V, 6V, 7V, 11V, 12V, FOV: 190 hfov, 130-160 vfov,  full 360, 8-32M pixel resolution
            \item \textbf{Impact}: Low light and adverse weather, algorithms are compute heavy, reliability issues
            \item \textbf{Advantages}: Dense information, passive sensor, inexpensive
        \end{itemize}
        \\
        \hline
        INS (Inertial Navigation Systems) &
        \begin{itemize}[nosep, leftmargin=*]
            \item INS provide reliable and continuous localization and motion data at meter level and sometimes cm level accuracy.
            \item INS uses accelerometers and gyroscopes to calculate the vehicle's position, orientation, and velocity.
            \item Allows filtering driven applications such as Multi-object tracking, forecasting, trajectory prediction.
        \end{itemize}
        \\
        \hline
        Other Sensors
        &
        \begin{itemize}[nosep, leftmargin=*]
            \item  High-speed Event based, Thermal, Hyper-spectral cameras, 
            \item Microphone or Acoustic Sensors and Cameras
            \item Driver monitoring Internal cameras, radars for heart-rate, gaze estimation
        \end{itemize}
        \\
        \hline
    \end{tabularx}
\end{table*}

% \begin{figure}[h]
% \includegraphics[width=\linewidth]{images/valeosensors.png}
% \caption{Sensor placement in a commercial vehicle and their field of view.}
% \label{fig:sensors}
% \end{figure}

\subsection{Dataset Requirements}
Large automotive datasets frequently are specified various parameters. There are mainly two types of parameters: Annotation specific, Environment specific. Annotation specific requirement parameters contain parameters like minimum number of unique objects, unique traffic lights or traffic signs. Annotation parameters also include annotation associated attributes occlusion, relevancy, lane association etc. Finally they specify which functions need to be covered: Eg. 3D Object detection, 3D Lane detection, TLR/TSR and Freespace. One can usually see these parameters specified in most of the AV datasets.

While environment specific parameters contain parameters under which dataset is created: example distance/volume collected in different countries across the word to represent different driving rules, metadata attributes for collection representing environmental conditions weather, lighting condition e.g., Night/Day , driving driving environment e.g., Urban/Highways.  Modern standards also demand "Chain of Causation" (CoC) annotations, where reasoning traces link observed scene evidence directly to driving decisions \cite{wang2025alpamayo}.

\subsection{Optimized dataset collection}
Data collection task takes the requirements as input and provides an optimal policy for collection. The optimality conditions include total logistic cost in terms of driving, data storage costs, annotation costs, model training costs, which are to be minimized. While data redundancy to handle loss of data in various conditions, model performance or accuracy, requirements to satisfy multiple industrial projects or deployments, require data collected to be maximized.

Dataset requirements gap completion is a sequential decision problem that defines the data collection policy that minimizes the amount of driven kilometres while maximizing the joint coverage of requirements conditions. It also ensures that the target model performance are reached. Authors have described such a system here \cite{moustafaconcept}. To reduce data storage costs data compression is performed on different sensors like cameras \cite{wang2023semantic}  and LiDAR \cite{roriz2024survey}.

\subsection{Automated data selection}
Datasets are usually separated into train and test subsets. Frequently a cross-validation subset is created to perform model selection and hyper-parameter fitting. Train subsets area constructed to ensure the model performance is satisfactory in different scenarios. Test subsets are constructed to evaluate performance rigorously across various scenarios or ODDs (operational design domains). This is to ensure broad evaluation of any AD stack model. Labelled datasets are split to ensure Train, Test subsets have proportional representation of labels and attributes (stratified sampling). Authors demonstrate \cite{almin2023navya3dseg} how dataset splitting is performed for point-cloud sequences for semantic segmentation task.

\begin{figure*}[h!]
    \centering
    \includegraphics[width=0.9\linewidth]{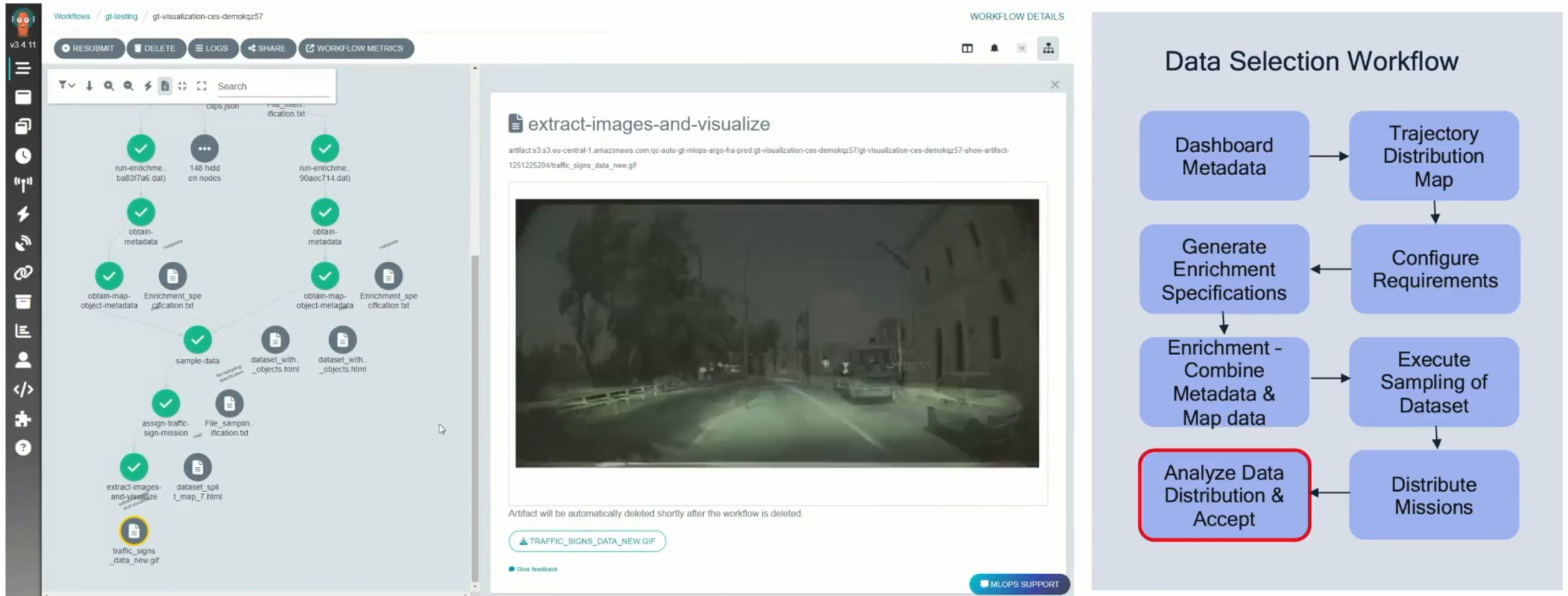}
    \caption{Automated data or file selection pipeline with various configurations to retrieve files that satisfy requirements, metadata attributes and filtering via OpenStreetMap and multimodal image-text embeddings.}
    \label{fig:data-selection-workflow}
\end{figure*}

Selection of scenes (tuple of multiple sensors) are performed to first ensure we satisfy dataset requirements, concurrently the samples that are selected are required to improve model performance on existing classes, specific metadata criteria or scenarios. Finally we also perform file selection to extract out of distribution samples between train and test sets or outlier samples. The automated file selection workflow, which is deployed in Argo, as shown in Fig\ref{fig:data-selection-workflow}.  Furthermore, the industry is adopting "automated MLOps pipelines" that use world models to synthesize rare "long-tail" scenarios, filling gaps in physical data collection through high-fidelity generative simulations \cite{zhao2025drivedreamer4d}.

\subsection{Dataset leakage}
Dataset leakage is a spurious relationship between the independent variables and the target variable that arises due data collection, dataset splitting, or other ways, that introduces information during training that is usually not available during inference. This usually leads to inflated estimates of model performance \cite{kaufman2012leakage}. Authors also advice on spatial, temporal and feature level separation to avoid leakage. When constructing large AV datasets there can potentially be overlaps between training and test datasets leading to leakage issues. Authors \cite{babud} address the issue of data leakage in automotive perception systems, particularly in object detection tasks. Method proposed leverages image similarity analysis (using perceptual hash (pHash)) to identify potential leakage between training and test sets. The method was validated through experiments on the Cirrus and KITTI datasets. Authors \cite{lilja2024localization} addresses the issue of data leakage in online mapping datasets, specifically nuScenes and Argoverse 2. The authors highlight that these datasets often revisit the same geographic locations across training, validation, and test sets, leading to inflated performance metrics. They propose geographically disjoint data splits to better evaluate the true performance of mapping methods in unseen environments. Experimental results show a significant drop in performance when proper data splits are used, revealing the impact of data leakage on current evaluation practices. The paper also reassesses prior design choices, finding that conclusions based on the original data splits may be misleading.  More recently, multimodal membership inference attacks have been proposed as a baseline pipeline to detect such contamination across vision and language modalities in advanced VLA models \cite{emelyanov2026fimmia}.

\subsection{Automated annotation quality check}
\begin{figure*}[h!]
    \centering
    \includegraphics[width=0.8\linewidth]{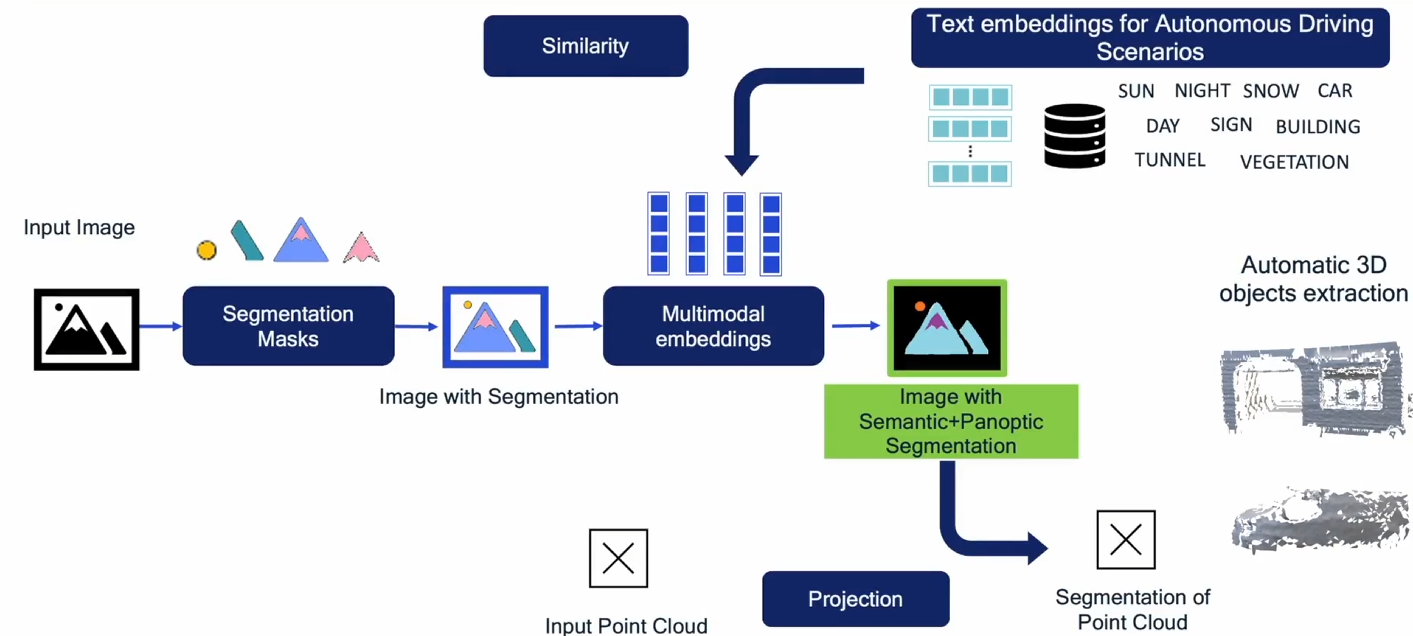}
    \caption{Automated annotation quality check model is a semantic segmentation pipeline based on SAM (Segment Anything Model) \cite{UltralyticsSAMDocs} and OpenClip \cite{OpenCLIP_2021}.}
    \label{fig:data-annotation-QC}
\end{figure*}

Offline autolabeling models are today used in large scale data annotation pipelines to reduce the cost of manual annotation \cite{ma2024zopp}. Though offline autolabeling models are not perfect and have errors in the outputs. Human based review of these errors are also slow process since model labelling errors are unpredictable. Automated quality check of annotations are automated methods to locate errors in the outputs of the annotation. We demonstrate our pipeline for automated review in figure \ref{fig:data-annotation-QC}. Automate error detection methods for object detectors have been studied for 2D methods in \cite{yatbaz2024run, tkachenko2023objectlab, ma2021delving}. The integration of foundation models such as SAM2 has significantly optimized this stage, allowing for semi-automatic mask generation that reduces manual annotation effort by up to 33-36\% while achieving accuracy nearly identical to manual ground truth \cite{some2026can}.

%%%%%%%%%%
\section{Dataset Safety Lifecycle}

\begin{figure}[h]
    \centering
    \includegraphics[width=0.45\textwidth]{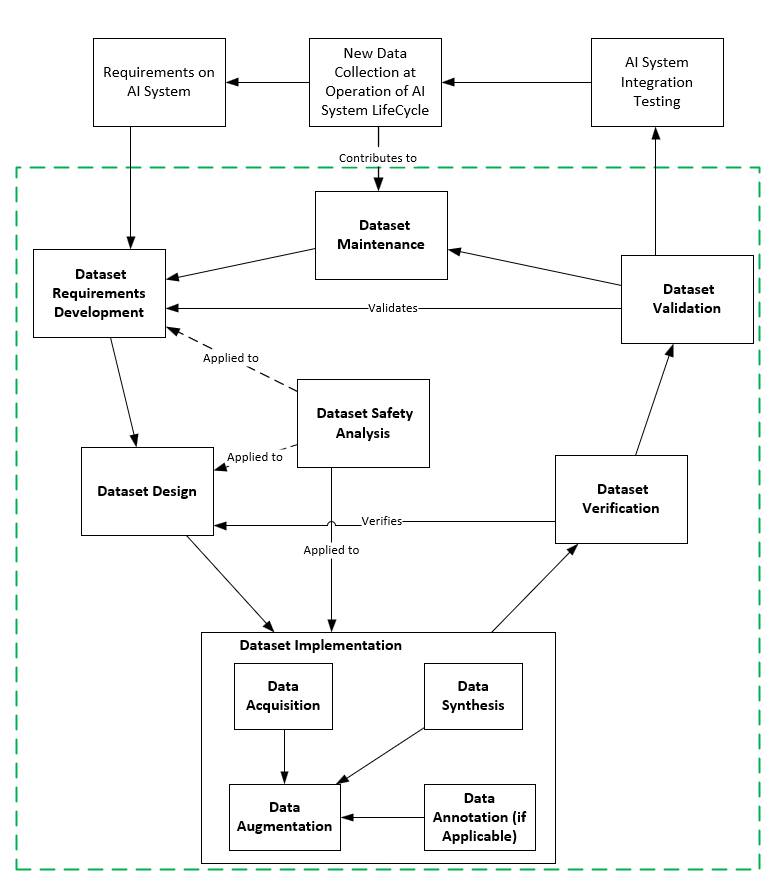}
    \caption{Dataset Lifecycle recommended by ISO/PAS 8800 \cite{ISO8800}}
    \label{fig:Lifecyle}
\end{figure}

The safety lifecycle management of datasets for autonomous driving systems plays a pivotal role in maintaining data quality, consistency, and reliability throughout the dataset’s usage period. A structured dataset lifecycle ensures rigorous compliance with safety standards and facilitates systematic improvements, accommodating evolving operational requirements and safety regulations.
ISO/PAS 8800 introduces a V-model based lifecycle which is presented in Fig.\ref{fig:Lifecyle}.

\subsection{Purpose of Dataset Lifecycle Management}

The primary goal of implementing a dataset lifecycle is to establish clear processes for developing, validating, verifying, and maintaining datasets. These processes ensure continuous improvement in dataset quality, promote compliance with regulatory frameworks, and support traceability throughout the entire lifecycle \cite{vyas2024key}.

\subsection{Use and Benefits of the V-model}

The V-model is a widely adopted approach in managing dataset lifecycles, especially in safety-critical domains such as autonomous driving. It provides a clear structure linking dataset requirements and design phases to subsequent implementation, verification, and validation phases. The model facilitates early detection of potential issues, reduces risks associated with data inaccuracies, and enhances communication among stakeholders by offering transparency and clarity at each development stage \cite{liu2024survey}. Additionally, using the V-model supports robust traceability by systematically linking dataset requirements to each corresponding implementation, verification, and validation activity. This structured traceability enables comprehensive documentation, facilitates regulatory compliance, and supports precise impact analysis of dataset modifications \cite{vyas2024key}

\subsection{Main Blocks in the Dataset Lifecycle}

\subsubsection{Dataset Safety Requirement Development}
Clearly defining dataset specifications based on safety standards, operational domain requirements, and AI model objectives. Requirements specify aspects such as data diversity, completeness, and quality thresholds essential for reliable autonomous driving models \cite{liu2024survey}.
\subsubsection{Dataset Design}
Planning data collection methodologies, generation strategies (physical, synthetic, augmented), data types, and core elements. This phase ensures alignment with the defined dataset requirements and includes metadata structuring for enhanced interpretability and traceability \cite{sarker2024comprehensive}.
\subsubsection{Dataset Implementation}
Executing data preparation processes, including physical collection, synthetic generation, augmentation, and meticulous labeling. This stage leverages advanced tools and automated systems to maintain high standards of accuracy, consistency, and efficiency \cite{liu2024survey}.
\subsubsection{Dataset Verification}
Conducting systematic evaluations to ensure the dataset meets predefined specifications. Verification involves checking data integrity, consistency, and compliance with standards through automated tools like TensorFlow Data Validation and rule-based verification methods \cite{vilas2025systematic}.
\subsubsection{Dataset Validation}
Assessing whether the dataset effectively fulfils the intended use in realistic scenarios. Validation incorporates performance metrics and scenario-based evaluations, utilizing both simulation and real-world testing to guarantee practical reliability \cite{vyas2024key}.

\subsubsection{Dataset Maintenance}
Continuous monitoring and updating of datasets throughout operational deployment to address emerging 
data drift, environmental changes, and evolving model requirements. Regular maintenance and 
revisions ensure long-term reliability and adaptability of datasets.

\section{Dataset Safety Requirements}

The development of dataset safety requirements begins with a thorough understanding of the AI system’s intended functionality. This includes identifying AI safety requirements and defining the input space, also known as the ODD. Each AI safety requirement and ODD aspect is explicitly mapped to corresponding dataset requirements, which helps define the specific scope of the dataset and the necessary level of detail to ensure the system operates safely and effectively within its intended context.

\subsection{Dataset safety properties}
\label{sec:Dataset safety properties}
In addition to the initial mapping of AI safety requirements and the ODD to dataset specifications, further dataset-specific safety requirements are systematically derived by applying dataset safety properties as outlined in \cite{ISO8800}. These properties such as accuracy, completeness, correctness, independence, integrity, representativeness, temporality, traceability, and verifiability \cite{ISO8800}, serve as foundational criteria for ensuring dataset reliability and safety. Depending on the application domain and operational context, additional relevant safety properties may also be identified and incorporated. These requirements are not static; rather, they emerge through iterative and comprehensive safety analyses conducted throughout the dataset lifecycle, including the phases of dataset requirement, dataset design, and implementation . This structured approach ensures that the dataset not only supports the AI system’s intended functionality but also aligns with rigorous safety and quality standards.
Detailed explanations of these properties are thoroughly described in ISO/PAS 8800 \cite{ISO8800}.\\

Given their significant relevance and frequent application in autonomous driving datasets, we elaborate on completeness and independence properties with explicit examples as follows:

\textbf{Completeness:}
Completeness ensures that all required data elements, including metadata, are comprehensively populated and effectively cover the defined input space, safety-critical scenarios, and plausible data perturbations \cite{ISO8800}. Illustrative examples of dataset incompleteness include scenarios where the dataset comprises only limited images of close-proximity obstacles, or lacks representations of nighttime conditions, despite nighttime operations being part of the intended operational domain.\\

\textbf{Independence of Datasets:}
Independence is another essential factor in ensuring dataset quality and reliability. Independence (also called dataset leakage) refers to the dataset’s ability to sufficiently prevent information leakage among datasets concerning their data sources and data collection methodologies \cite{ISO8800}. Common violations of independence include separating sequential video frames across different datasets (e.g., training and test datasets), or utilizing data exclusively sourced from identical environments (such as the same city streets, weather conditions, and traffic scenarios).\\

\subsection{Safety Analysis throughout the Dataset Lifecycle}
To ensure comprehensive and robust coverage of dataset requirements, systematic safety analyses are conducted across main phases of the dataset lifecycle, including dataset requirement, dataset design, and implementation. These analyses are instrumental in identifying potential deficiencies or vulnerabilities that may compromise the dataset’s ability to support the AI system’s safety and performance objectives. In response to the insights gained from these evaluations, additional dataset requirements are formulated to address identified gaps, thereby reinforcing the dataset’s alignment with the system’s operational and safety expectations. The methodology and outcomes of the safety analysis are discussed in detail in Section VII.

\section{Dataset Design}

The design of datasets for autonomous driving applications must strictly adhere to predefined dataset requirements to ensure comprehensive representation and reliability. Effective dataset design involves systematically collecting, generating, augmenting, and maintaining data elements aligned with safety standards and operational needs.

\subsection{Data Elements and Generation Methods:}

Datasets typically incorporate three categories of data elements: physically collected, synthetically generated, and augmented data. Physical data collection involves capturing real-world driving scenarios using vehicle-mounted sensors such as LiDAR, radar, cameras, GPS, and inertial measurement units (IMUs). For example, the KITTI dataset includes extensive physically collected data from sensor arrays capturing varied driving environments\cite{geiger2012we}.\\

Synthetic data generation, an alternative approach, uses simulation tools to create scenarios that are difficult, expensive, or unsafe to replicate physically. CARLA and LGSVL simulators, for instance, provide virtual environments capable of generating realistic driving scenarios that include rare edge-cases and hazardous conditions \cite{dosovitskiy2017carla, rong2020lgsvl}.\\

Data augmentation methods further enhance datasets by artificially expanding the dataset through transformations applied to existing data points. Techniques such as rotation, scaling, brightness adjustment, and noise addition help improve model robustness against variations in operational conditions without extensive additional data collection \cite{shorten2019survey}.

\subsection{Core Data Elements:}

Core data elements essential to autonomous driving datasets include sensor data streams, temporal synchronization between sensors, object detection and tracking, semantic segmentation, and scenarios capturing dynamic and static environmental interactions. Specifically, the Berkeley DeepDrive dataset highlights core elements such as object annotations and dense semantic segmentation, emphasizing their role in accurate perception model training \cite{yu2020bdd100k}.

\subsection{Metadata and Ground Truth:}

Metadata encompasses supplementary information vital for interpreting the primary dataset, including environmental conditions (weather, lighting), geographic information (GPS coordinates), and sensor calibration details. Crucially, ground truth annotations, including object class labels, bounding boxes, segmentation masks, and velocity vectors, provide foundational truth against which AI models are evaluated. The nuScenes dataset exemplifies comprehensive metadata integration, offering rich annotations for dynamic objects across multiple sensor modalities \cite{caesar2020nuscenes}.

\subsection{Dataset Operations:}

Effective dataset design also includes performing systematic operations to maintain quality and compliance with safety and privacy standards. Operations such as filtering irrelevant or redundant data ensure dataset relevance and computational efficiency. Dimensionality reduction techniques, like Principal Component Analysis (PCA) or t-distributed Stochastic Neighbour Embedding (t-SNE), help in optimizing dataset representation and reducing computational load \cite{van2008visualizing}.

Moreover, data privacy mandates necessitate implementing robust de-identification processes, such as anonymization or pseudonymization of identifiable markers, to protect individual privacy according to regulations like GDPR and HIPAA \cite{hintze2018comparing}.

\subsection{Monitoring Mechanisms for Distribution Shift:}

Dataset design must incorporate mechanisms for real-time monitoring of distribution shifts during operational deployment.
Real-time shift detection addresses distributional changes without immediate ground-truth labels by monitoring proxy signals that are available at inference time.  In practice, this includes \emph{(i) feature distribution divergence}, where distances/divergences are computed between reference and live feature embeddings (e.g., deep representation distances) to flag drift efficiently in streaming settings \cite{greco2025unsupervised},  \emph{(ii) confidence-based uncertainty}, where changes in predictive confidence/uncertainty act as early warning indicators of mismatch between training and deployment conditions \cite{he2026survey},  and \emph{(iii) out-of-distribution (OOD) detectors}, which explicitly monitor whether incoming inputs deviate from the training distribution as part of runtime assurance \cite{pitale2024inherent}. 

Offline performance monitoring complements these proxies by using delayed or curated labels to quantify true task performance.  Standard accuracy and precision/recall (and their variants) provide label-based estimates of correctness and class-specific error tradeoffs,  while scenario-based metrics stratify performance by operationally meaningful conditions (e.g., scenario families in simulation or structured test suites) to localize degradation to specific regimes.
Techniques such as statistical distance measures (e.g., Kullback-Leibler divergence, Jensen-Shannon divergence) and performance metrics monitoring (accuracy, precision, recall) provide essential feedback for detecting data drift \cite{moreno2012unifying}. Continuous data collection strategies help capture newly encountered scenarios, enabling iterative dataset revision and updating to maintain model effectiveness under evolving operational conditions.

%%%%%%%%%%%%%
\subsection{Training, Test, and Development Datasets}
In autonomous driving AI model development, datasets are commonly partitioned into three distinct subsets: training, test, and development (validation) datasets. Each serves a specific purpose and must be carefully managed to prevent data leakage and overfitting.\\

\textbf{Training Dataset:} The training dataset is the largest subset and is used exclusively to train the AI model. It should cover a broad spectrum of scenarios and include representative examples of typical driving conditions and rare edge cases. Adequate data augmentation practices can further increase the effective size and diversity of the training set \cite{shorten2019survey}.\\

\textbf{Development (Validation) Dataset:}
The development dataset, also known as the validation set, is used to fine-tune hyperparameters and evaluate model performance during training iterations. It should be distinct from the training set but representative enough to reflect real-world driving conditions. This dataset helps identify overfitting and guides decisions about model adjustments.\\

\textbf{Test Dataset:} The test dataset is used exclusively to evaluate the final performance of the trained model. It must be entirely separate from both the training and validation datasets to provide an unbiased assessment of the model’s generalization capabilities. The test set should ideally include challenging scenarios, representing diverse driving conditions and edge cases the model might encounter in operation \cite{geiger2012we}.\\

\textbf{Preventing Overfitting:}
Overfitting occurs when a model learns specific patterns from the training data that do not generalize well to unseen data. To prevent overfitting, it is crucial to ensure that datasets are partitioned without overlap, using methods like stratified random sampling or temporal splits. Techniques such as cross-validation, early stopping, and regularization (e.g., dropout or weight decay) are also commonly employed to mitigate overfitting \cite{goodfellow2016deep}.\\

Careful management and systematic separation of datasets ensure the model’s robustness and reliability in real-world deployment.

\section{Dataset Implementation}
Dataset implementation encompasses structured processes to ensure the preparation, labeling, and management of data tailored specifically for AI development. In the following, the core operations in dataset implementation are explained.

\subsection{Dataset Preparation}

Dataset preparation involves several key steps: physical data collection, synthetic data generation, and augmentation, all aligned with the dataset design discussed in the previous section. Physically collected data typically rely on vehicle-mounted sensors such as LiDAR, radar, and cameras. Recent datasets like the Waymo Open Dataset showcase high-quality, diverse data capturing complex urban scenarios \cite{sun2020scalability}.

Data preparation also includes thorough cleaning, normalization, and verification to ensure quality and consistency. Tools like TensorFlow Data Validation (TFDV) and Apache Spark support scalable data processing. Automated anomaly detection methods, as demonstrated by Hu et al. \cite{hu2022processing}, significantly reduce manual effort and enhance reliability.

\subsection{Defining Processes, Methods, and Tools for Dataset Labeling}

Labeling datasets accurately is vital for supervised learning applications. Manual annotation processes, though accurate, are labor-intensive and costly. Recent research by Lee et al. \cite{lee2021semi}introduced semi-automated labeling frameworks, utilizing machine learning-assisted annotation to improve labeling speed and consistency. Commercial tools like Amazon SageMaker Ground Truth and Scale AI leverage AI-assisted annotation techniques, drastically enhancing labeling efficiency. Moreover, these platforms incorporate rigorous quality control mechanisms, ensuring label accuracy and reliability.

\subsection{ Labeling the Dataset}

The labeling phase involves annotating sensor data with precise metadata such as bounding boxes, segmentation masks, velocity vectors, and object class labels. Recent works, like the nuScenes dataset, exemplify comprehensive labeling approaches across multiple sensor modalities, ensuring high accuracy and detailed annotations \cite{caesar2020nuscenes}. Additionally, advances in 3D labeling techniques, as demonstrated in the KITTI benchmark suite \cite{geiger2012we}, significantly enhance spatial accuracy critical for depth perception tasks. Continuous development in labeling methodologies ensures datasets remain precise, consistent, and scalable for evolving autonomous driving technologies.

\subsection{Dataset Compression}

Dataset compression in ADAS development addresses storage constraints, reduces data transfer times, and improves training efficiency across large-scale multimodal data, including camera images, LiDAR point clouds, and radar signals. Both lossless and lossy compression methods can be applied: lossless compression preserves all original information, ensuring exact recovery of categorical and textual data; lossy compression reduces size by discarding redundant or imperceptible details, which can be acceptable if essential features for model learning remain intact \cite{underwood2024understanding}.\\

To preserve core data elements, compression parameters are chosen to maintain critical content while minimizing quality loss. Error-bounded compression (e.g., SZ \cite{SZCompressorOverview} or ZFP \cite{ZfpWebsite}) ensures deviations from original data stay within a defined threshold, maintaining visual and statistical fidelity \cite{di2016fast}. Compression must be applied consistently across training, validation, and test datasets to avoid distribution shifts \cite{jiang2025iterative}. ISO/PAS 8800’s dataset safety principles reinforce that compression should not degrade safety-critical information, especially in safety-relevant scenarios \cite{ISO8800}.\\

Validating compression safety relies primarily on model performance benchmarking. Accuracy, F1-score, and other task-relevant metrics are compared for models trained and validated on original versus compressed datasets. If performance differences remain negligible, the compression is considered safe. Studies show that image datasets can be significantly compressed without measurable accuracy loss, and in some cases, moderate compression even improves generalization by removing noise \cite{zhou2024deep}.

Additional validation includes side-by-side predictions, statistical distribution comparisons, and targeted scenario testing to detect degradation in rare but safety-critical cases \cite{underwood2024understanding}. By combining careful method selection, consistent application across dataset partitions, and rigorous performance evaluation, dataset compression can be safely integrated into ADAS workflows by optimizing storage and computation without compromising the integrity and utility of data for training and validation.

\section{Safety Analysis Methods}
\label{sec:Methods}
Ensuring dataset safety for autonomous driving (AD) AI models is critical. Dataset issues directly affect AI reliability, necessitating rigorous safety analyses. ISO/PAS 8800 standard recommend to conduct safety analysis on dataset requirements, dataset design, and dataset implementation \cite{ISO8800}. This paper discusses four primary methods—HAZOP, FTA, FMEA/PFMEA, and STPA—for assessing dataset requirements, design, and implementation stages.
\subsection{Hazard and Operability Study (HAZOP)}
HAZOP systematically identifies potential hazards using guidewords like "No," "Less," and "Wrong" to detect dataset deviations (e.g., missing nighttime pedestrian data)\cite{lawley1974operability, qi2022hierarchical}. It is particularly useful during dataset requirement definition and design, identifying coverage gaps and labeling errors.

\textbf{Advantages:}
\begin{itemize}
  \item Structured, creative hazard discovery.
  \item Adaptable to ML-specific concerns.
\end{itemize}

\textbf{Limitations:}
\begin{itemize}
  \item Labor-intensive, reliant on expert judgment.
  \item Qualitative; needs adaptation for ML applications.
\end{itemize}

\subsection{Fault Tree Analysis (FTA)}
FTA is a deductive method tracing back from a hazard (top event) to root causes like dataset deficiencies (e.g., lack of pedestrian examples) \cite{aoki2020dataset}. FTA effectively links known dataset issues to specific system failures, aiding in requirements validation and risk quantification.

\textbf{Advantages:}
\begin{itemize}
  \item Logical cause-effect structure.
  \item Supports risk quantification.
\end{itemize}

\textbf{Limitations:}
\begin{itemize}
  \item Requires predefined hazards.
  \item Complexity with extensive data scenarios.
\end{itemize}

\subsection{Failure Mode and Effects Analysis (FMEA/PFMEA)}
FMEA identifies possible dataset component failures and their system impacts (e.g., labeling errors, scenario omissions) through systematic bottom-up analysis \cite{schmitt2025introducing}. PFMEA focuses on dataset processes, ensuring comprehensive quality control across data collection, annotation, and validation phases.

\textbf{Advantages:}
\begin{itemize}
  \item Granular, actionable insights.
  \item Prioritizes issues effectively.
\end{itemize}

\textbf{Limitations:}
\begin{itemize}
  \item Potentially exhaustive and resource-intensive.
  \item May miss interactions between failures.
\end{itemize}

\subsection{System-Theoretic Process Analysis (STPA)}
STPA identifies unsafe control actions within system interactions, including data-driven ML model training processes \cite{qi2023stpa}. It captures emergent hazards from complex interactions (e.g., dataset distribution shifts causing model failures) and generates comprehensive safety constraints.

\textbf{Advantages:}
\begin{itemize}
  \item Captures systemic, interaction-based hazards.
  \item Derives broad safety constraints.
\end{itemize}

\textbf{Limitations:}
\begin{itemize}
  \item Requires expertise; less intuitive.
  \item Generates many scenarios; prioritization is challenging.
\end{itemize}

Table \ref{tab:comparison} summarizes the comparison of these safety methods regarding their focus, dataset lifecycle applicability, strengths, and limitations.

\begin{table*}[h!]
\centering
\caption{Comparison of Dataset Safety Analysis Methods}
\label{tab:comparison}
\small
\begin{tabular}{@{}p{2.3cm}p{3.3cm}p{2.5cm}p{3.8cm}p{3.8cm}@{}}
\toprule
Approach & Technique & Dataset Stage & Strengths & Limitations \\
\midrule
HAZOP & Guideword-driven deviations & Requirements, Design & Systematic, broad hazard discovery & Labor-intensive, qualitative \\
FTA & Deductive logic tree & Design, Implementation & Logical structure, risk quantification & Needs predefined hazards, complexity \\
FMEA/PFMEA & Inductive failure analysis & Design, Implementation & Granular detail, actionable, prioritizes well & Extensive, misses interactions \\
STPA & Unsafe control action scenarios & Requirements, Design & Systemic interaction-based hazards & Expert-dependent, many scenarios \\
\bottomrule
\end{tabular}
\end{table*}

These methods complementarily enhance dataset safety in autonomous driving. HAZOP and STPA effectively discover broad and systemic hazards early. FTA and FMEA provide structured verification and granular quality assurance. An integrated approach leveraging all four ensures robust dataset safety across the AI lifecycle.

%%%%%%%%%%%%%%%%%%%%%%
\section{Dataset Requirement Verification \& Validation}

Verification and validation are fundamental activities for ensuring the safety, reliability, and trustworthiness of AI-based systems in road vehicles. According to ISO/PAS~8800, \emph{verification} is defined as the confirmation, through the provision of objective evidence, that specified requirements have been fulfilled. In contrast, \emph{validation} is defined as the confirmation, through objective evidence, that the requirements for a specific intended use or application have been satisfied. \\
This section first reviews representative dataset verification methods reported in the literature. Subsequently, a case study is presented to illustrate how these methods can be practically applied in an automotive AI context. Finally, existing dataset verification and validation (V\&V) approaches are systematically compared to highlight their scope, assumptions, and limitations.

\subsection{Dataset Verification Methods}
\label{sec:dataset-verification}

In the context of AI systems, it is important to verify that the AI system fulfills its safety requirements. Validation should ensure that the safety requirements allocated to the AI system are met when it is integrated into the encompassing system. Safety analysis, performed using the methods mentioned in Section~\ref{sec:Methods} can help identify additional verification criteria to be applied during dataset validation. \\
%%%%%%%%%%%%%
\begin{table}[h!]
\scriptsize
\caption{Verification and Validation Test Methods and Use Cases}
\centering
\begin{tabular}{|p{4cm}|p{4cm}|}
\hline
\textbf{Test Methods} & \textbf{Use Case} \\ \hline
Analysis of requirements & Selecting key performance indicators (KPIs) for verification and validation activities \\ \hline
Generation and analysis of equivalence classes & Generating comprehensive test sets for pre- and post-processing algorithms \\ \hline
Error guessing based on knowledge or experience of human as well as knowledge automatically derived from an algorithm algorithm & Identifying unknown edge cases for testing based on knowledge or experience \\ \hline
Analysis of boundary values & Creating complete test sets for pre- and post-processing algorithms \\ \hline
\end{tabular}
\label{table:VandV}
\vspace{0.5cm}
\end{table}
%%%%%%%%%%%%%%%%%

These verification criteria can be used as a measure to avoid or prevent certain potential failure modes.
Table \ref{table:VandV}  talks about the test methods which can be used for verification and validation of dataset. These test methods are explained in detail with one requirement each for each test method in the following:\\
%%%%%%%%%%%%
\subsubsection{Test Method - Analysis of requirements}
The purpose of this method is to verify the completeness, correctness, consistency and testability of the requirement. Analysing requirements lays the groundwork for creating effective test cases. By thoroughly understanding what the system should do, testers can identify the appropriate inputs, predict the expected results, and focus on verifying key functionalities. Analysis of dataset requirement starts by collecting reviewers comments about the listed requirements to decide whether the requirements are unambiguous, comprehensible, atomic(singular), internally consistent, feasible, verifiable, etc., as per ISO 26262-8.
We have considered DSR-01 (DSR - dataset Requirement) as an example for 'analysis of requirements' test method.\\

\begin{itemize}
    \item DSR 01: The real world test dataset shall include at least “2200” number of frames.\\
\end{itemize}
Verification reviewers also check if this DSR has upstream traceability to ASR (AI System Requirement) and DSDR (Dataset Design Requirement) as per ISO PAS 8800.

\subsubsection{Test Method - Generation and analysis of equivalence classes}
\label{sec:equivalence classes}
Analyzing Equivalence Classes is a crucial aspect of software testing, as it promotes thorough test coverage while reducing unnecessary or repetitive test cases. Equivalence partitioning is a testing technique in which input data is divided into distinct classes, each representing a range of values expected to trigger similar system responses, thus streamlining the testing process.
 For example, the following DSR 02 is considered for this test method.\\ 

\begin{itemize}
    \item DSR 02 - The training and validation datasets for semantic segmentation must consist of at least 10,000 images. Additionally, every image in these datasets must be fully annotated to ensure data integrity and usefulness during model training and evaluation. \\
\end{itemize}

Here, we can define the equivalence classes as mentioned below with corresponding result of a test case. Table \ref{table:equivalence_classes} summarizes this test method with test cases (TC) and expected results. 
 The evaluation approach involves generating and analyzing equivalence classes based on both the total number of images and their annotation status. The pass/fail criteria are clearly defined:

\begin{itemize}
    \item
Pass: The dataset passes if the total image count is greater than or equal to 10,000 and all images are annotated. Formally, Pass $\Leftrightarrow$ (Total images $\geq$ 10,000) AND (Annotated images $=$ Total images).
\item
Fail: The dataset fails if either the total image count is $<$ 10,000 or not all images are annotated. Formally, Fail $\Leftrightarrow$ Otherwise. \\
\end{itemize}

Table \ref{table:equivalence_classes} presents the generation and analysis of equivalence classes method, outlining various test cases and their expected outcomes. Valid equivalence classes, as per DSR 02, include datasets with 11,000, 12,500, and 15,000 images, where all images are annotated. These cases meet both the minimum count and annotation requirements, resulting in a "Pass" outcome for test cases TC02.1, TC02.2, and TC02.3.
Invalid equivalence classes include datasets with 9,000, 7,500, and 5,000 images—even though all are annotated, the total count is below the required threshold. These are represented by test cases TC02.4, TC02.5, and TC02.6, each resulting in "Fail." Additionally, test case TC02.7 demonstrates failure due to incomplete annotation: although the dataset contains 11,000 images, only 9,500 are annotated.

\begin{table}[!t]
\centering
\caption{Test Cases for Image Annotation Equivalence Classes}
\label{table:equivalence_classes}
\scriptsize
\begin{tabular}
{|p{1cm}|p{1.5cm}|p{1.5cm}|p{3cm}|}
\hline
\textbf{Test Case number} & \textbf{Total number of Images} & \textbf {Total number of annotated Images} & \textbf{Result of Pass/Fail criteria} \\
\hline
TC 02.1 & 11,000 & 11,000 & Pass \\
TC 02.2 & 12,500 & 12,500 & Pass \\
TC 02.3 & 15,000 & 15,000 & Pass \\
TC 02.4 & 9,000 & 9,000 & Fail (below minimum count) \\
TC 02.5 & 7,500 & 7,500 & Fail (below minimum count) \\
TC 02.6 & 5,000 & 5,000 & Fail (below minimum count) \\
TC 02.7 & 11,000 & Only 9,500 & Fail (incomplete annotation) \\
\hline
\end{tabular}
\vspace{0.5cm}
\end{table} 

\subsubsection{Test Method - Error guessing based on prior knowledge} 
% or experience can be suited to identify yet unknown edge cases for testing}
Error guessing is a test design technique in which domain experts leverage prior experience and system knowledge to identify potential failure modes that may not be explicitly captured in formal specifications. Rather than relying exclusively on requirement‑driven test derivation, this approach anticipates system behavior under edge cases, unexpected inputs, and stress conditions. By drawing on knowledge of similar systems, historical defects, and known failure mechanisms, engineers can design targeted test cases aimed at exposing defects that are unlikely to emerge through conventional specification‑based testing alone.\\
In the context of this work, the error‑guessing method is applied using domain knowledge to address a representative failure mode, namely the misclassification of shadows as obstacles, which is a well‑known source of perceptual error in vision‑based automotive systems.\\

\textbf{Error guessing steps}:
\begin{enumerate}[label=\alph*), leftmargin=*]
    \item Create synthetic data to simulate the above condition, i.e., misclassification of shadows as obstacles.
    \item Use real-world datasets containing known edge cases, i.e., misclassification of shadows as obstacles.
\end{enumerate}

\subsubsection{Test Method - Analysis of boundary values}
\begingroup

We will consider the DSR 03 as an example.
\begin{itemize}
    \item DSR 03 - For the pedestrian semantic class, the dataset shall include annotated samples whose level of occlusion does not exceed the maximum occlusion threshold of 70\%. \\
\end{itemize}
The pass/fail criteria is defined as follows.\\
\begin{itemize}
    \item
\textbf{Pass Criterion:} A test case shall be considered \emph{Pass} if the measured occlusion level of the pedestrian annotation is less than or equal to 70\% and the occlusion level is derived and annotated in accordance with the defined annotation methodology.
\item
\textbf{Fail Criterion:} A test case shall be considered \emph{Fail} if the measured occlusion level of the pedestrian annotation exceeds 70\%, or if the occlusion level is incorrectly determined, inconsistently annotated, or otherwise invalid, even when the measured value is equal to the threshold.\\
\end{itemize}

Table \ref{table:boundary_values} summarizes the boundary value test cases used to verify the pedestrian occlusion requirement. The test cases evaluate occlusion levels at the defined upper boundary (70\%), as well as values immediately below and above this threshold. The results demonstrate that annotated samples with occlusion levels less than or equal to 70\% satisfy the requirement, while samples exceeding the threshold or containing invalid occlusion annotations are correctly identified as failures. This confirms that the dataset verification adequately covers the boundary conditions associated with pedestrian occlusion. \\

\begin{table}[!t]
\centering
\caption{Boundary Value Test Cases for Image Annotation}
\label{table:boundary_values}
\scriptsize
\begin{tabular}
{|p{1cm}|p{1.5cm}|p{1.5cm}|p{3cm}|}
\hline
\textbf{Test Case number} &
\textbf{Measured occlusion level (\%)} &
\textbf{Maximum allowed (\%)} &
\textbf{Result of Pass/Fail criteria} \\
\hline
TC 03.1 & 70.0 & 70 & Pass \\
\hline
TC 03.2 & 69.9 & 70 & Pass \\
\hline
TC 03.3 & 69.8 & 70 & Pass \\
\hline
TC 03.4 & 70.1 & 70 & Fail (exceeds occlusion threshold) \\
\hline
TC 03.5 & 70.2 & 70 & Fail (exceeds occlusion threshold) \\
\hline
TC 03.6 & 70.0 (incorrectly annotated) & 70 & Fail (invalid boundary annotation) \\
\hline
\end{tabular}
\end{table}
\endgroup
\begin{table*}[h!]
\centering
\caption{Overview of Datasets, Simulation Tools, and Validation Frameworks with Data Characteristics, ODD Coverage, and Safety Properties}
\label{tab:dataset_overview_wide_nolinks}
\scriptsize
\setlength{\tabcolsep}{6pt}
\renewcommand{\arraystretch}{1.15}
\begin{tabular}{p{3.6cm} p{4cm} p{4cm} p{4cm}}
\hline
\textbf{Paper / Dataset} &
\textbf{Type of Data Used} &
\textbf{ODD / Scenario Coverage} &
\textbf{Safety Properties Addressed} \\
\hline

KITTI \cite{Geiger2012KITTI} &
Real-world multimodal (stereo camera imagery, LiDAR, localization metadata) &
Mid-size city, rural roads, and highways; limited geographic diversity &
\textit{Explicit:} Integrity, Correctness, Traceability, Verifiability \newline
\textit{Partial/Implicit:} Accuracy, Independence, Completeness, Representativeness \\

nuScenes \cite{caesar2020nuscenes} &
Real-world multimodal (surround-view cameras, radar, LiDAR, vehicle state/pose) &
1000 urban scenes in Boston and Singapore; diverse traffic scenarios &
\textit{Explicit:} Integrity, Correctness, Traceability, Verifiability \newline
\textit{Partial/Implicit:} Accuracy, Temporality, Completeness, Representativeness \\

Waymo Open Dataset \cite{Sun2020Waymo} &
Real-world multimodal (multi-view cameras, LiDAR, calibration and pose metadata) &
Urban and suburban environments across multiple regions &
\textit{Explicit:} Integrity, Correctness, Traceability, Verifiability \newline
\textit{Partial/Implicit:} Accuracy, Temporality, Completeness, Representativeness \\

CARLA Simulator \cite{dosovitskiy2017carla} &
Synthetic simulation data (configurable sensor suite, privileged ground truth) &
Fully configurable ODD including weather, lighting, traffic, maps, and sensors &
\textit{Explicit:} Traceability, Verifiability \newline
\textit{Partial/Implicit:} Integrity, Temporality, Completeness, Representativeness \\

LGSVL / SVL Simulator \cite{rong2020lgsvl} &
Synthetic high-fidelity simulation (stack-integrated, configurable sensors) &
Configurable ODD via digital twins, maps, sensors, and controllable objects &
\textit{Explicit:} Traceability, Verifiability \newline
\textit{Partial/Implicit:} Integrity, Temporality, Completeness, Representativeness \\

DAIR-V2X \cite{Yu2022DAIRV2X,DAIRV2X_Website} &
Real-world cooperative (vehicle-side and infrastructure-side camera and LiDAR) &
Urban intersections; cooperative V2I perception; day/night and weather variation &
\textit{Explicit:} Integrity, Correctness, Traceability, Verifiability \newline
\textit{Partial/Implicit:} Accuracy, Temporality, Completeness, Representativeness \\

Anomaly Dataset Survey \cite{Bogdoll2023AnomalySurvey} &
Mixed datasets (real anomalies, injected anomalies, fully synthetic scenes) &
Rare events, anomalies, and out-of-distribution scenarios &
\textit{Explicit:} Traceability, Verifiability, Completeness \newline
\textit{Partial/Implicit:} Temporality, Representativeness \\

Leakage Study \cite{Lilja2023DataLeakage} &
Real-world mapping datasets (geographic overlap and leakage analysis) &
Unseen-location evaluation; addresses geographic revisits across splits &
\textit{Explicit:} Independence, Traceability, Verifiability \\

TFDV \cite{TFDV_Documentation} &
Dataset validation tooling (schema, statistics, drift and skew detection) &
Applicable across evolving operational conditions via windowed analysis &
\textit{Explicit:} Integrity, Temporality, Traceability, Verifiability \\

\hline
\end{tabular}
\end{table*}

\subsection{Worked Case Study: Dataset Safety Assurance for ADAS Perception} To illustrate the practical application of the test methods defined in Table \ref{table:VandV}, we present a worked case study centered on the semantic segmentation requirements (DSR 02,03). This study follows the ISO/PAS 8800 ``Dataset V-Model,'' which defines data as a safety-critical artifact requiring explicit versioning, traceability, and representativeness.\\

\textbf{1. Verification via Requirement Analysis and Traceability:}
Verification initiates with a structured analysis of DSR 02 (10,000 fully annotated images'') to confirm it is atomic and unambiguous. Reviewers verify upstream traceability to AI System Requirements (ASR), ensuring the 10,000-image volume is statistically adequate to characterize the target ODD. This phase applies the Changing Anything Changes Everything'' principle, where the impact of training mini-batches on system-level predictability is formally documented \cite{ISO8800}.\\
\textbf{2. Conformance Verification through Equivalence and Boundary Testing:} The dataset is processed using the test suites in Table \ref{table:equivalence_classes} and Table \ref{table:boundary_values}:
\begin{itemize}
\item \textit{Process Integrity Verification:} In TC 02.7 (11,000 images, 9,500 annotated), the validation tool triggers a ``fail'' status. This identifies a process insufficiency where 1,500 frames missed the annotation pipeline, preventing functional degradation in the trained model.
\item \textit{Boundary Robustness Verification:} Testing at maximum 70\% threshold (TC 03.1 vs TC 03.6) validates that the data ingestion pipeline correctly enforces maximum safety limits, preventing off-by-0.1 errors.\end{itemize}
\textbf{3. Suitability Validation via Error Guessing and World Models:}Validation confirms the dataset is suitable for its intended use.
\begin{itemize}\item \textit{Hazard Analysis:} ``Error Guessing'' for shadows identifies failure pathways where pedestrians are occluded by high-contrast lighting.\\
\item \textit{Closing the Safety Gap:} To address identified insufficiencies, generative world models or simulators (e.g., CARLA) are used to synthesize hazardous shadow scenarios \cite{dosovitskiy2017carla, hu2023gaia, wang2024driving}. These are injected into the test subset, providing objective evidence that the AI system fulfills safety requirements under extreme ODD conditions \cite{dauner2024navsim, yang2025drivearena}.\end{itemize}This structured approach transforms dataset management into a verified safety engineering discipline aligned with emerging automotive standards.
%%%%%%%%%%%%%%%%%%%%
\begin{table*}[t]
\centering
\caption{Assessment of Dataset and Tool Support Across Quality and Assurance Properties}
\label{tab:dataset_assessment_wide}
\scriptsize
\setlength{\tabcolsep}{4pt}
\begin{tabular}{lccccccccc}
\hline
\textbf{Reference} & \textbf{Acc} & \textbf{Comp} & \textbf{Corr} & \textbf{Ind} & \textbf{Int} & \textbf{Repr} & \textbf{Temp} & \textbf{Trac} & \textbf{Verif} \\
\hline
KITTI (Geiger et al. \cite{Geiger2012KITTI}) & P & P & E & P & E & P & -- & E & E \\
nuScenes (Caesar et al. \cite{caesar2020nuscenes}) & P & P & E & -- & E & P & P & E & E \\
Waymo Open Dataset (Sun et al. \cite{Sun2020Waymo}) & P & P & E & -- & E & P & P & E & E \\
CARLA Simulator (Dosovitskiy et al. \cite{dosovitskiy2017carla}) & N/A & P & P & N/A & P & P & P & E & E \\
LGSVL / SVL Simulator (Rong et al. \cite{rong2020lgsvl}) & N/A & P & P & N/A & P & P & P & E & E \\
DAIR-V2X (Yu et al. \cite{Yu2022DAIRV2X,DAIRV2X_Website}) & P & P & E & -- & E & P & P & E & E \\
Anomaly Dataset Survey (Bogdoll et al. \cite{Bogdoll2023AnomalySurvey}) & N/A & E & N/A & N/A & N/A & P & P & E & E \\
Leakage Study (Lilja et al. \cite{Lilja2023DataLeakage}) & N/A & N/A & N/A & E & N/A & N/A & N/A & E & E \\
TensorFlow Data Validation (TFDV \cite{TFDV_Documentation}) & N/A & N/A & N/A & N/A & E & N/A & E & E & E \\
\hline
\multicolumn{10}{l}{\textbf{Legend:} E = explicit evidence; P = partial/implicit evidence; -- = not evidenced; N/A = not applicable} \\
\multicolumn{10}{p{0.98\textwidth}}{\textbf{Abbreviations \& Definitions:}
\textbf{Acc} = Accuracy (documented annotation specifications, QA processes, or quantified label error);
\textbf{Comp} = Completeness (explicit coverage targets or metadata completeness checks);
\textbf{Corr} = Correctness/Fidelity (evidence that data correspond to the intended real-world phenomenon);
\textbf{Ind} = Independence (explicit split policies and leakage or overlap analysis);
\textbf{Int} = Integrity (controls against data corruption or processing errors);
\textbf{Repr} = Representativeness (analysis of distributional bias relative to the intended ODD);
\textbf{Temp} = Temporality (consideration of dataset aging, drift, or refresh/versioning policies);
\textbf{Trac} = Traceability (reproducible artifacts enabling reconstruction of dataset usage);
\textbf{Verif} = Verifiability (objective evaluation procedures and machine-checkable evidence).} \\
\end{tabular}
\end{table*}
%%%%%%%%%%%%%%%%

\subsection{Comparison of different testing methods in dataset V\&V}

ISO/PAS 8800 mentions that dataset-related safety properties and dataset related insufficiencies should be taken into consideration while dealing with safety critical applications which use AI. This ensures datasets are free from errors, biases and inconsistencies. A study of scholarly articles was undertaken to explore the data related safety properties used, verification and validation methodologies used in autonomous driving with different types of datasets such as Real-world data, synthetic data, Collaborative perception data etc.\cite{Geiger2012KITTI} \cite{caesar2020nuscenes}  \cite{Sun2020Waymo} \cite{dosovitskiy2017carla} \cite{rong2020lgsvl}  \cite{OSSDC_SVL_Docs} \cite{Yu2022DAIRV2X,DAIRV2X_Website} \cite{Bogdoll2023AnomalySurvey} \cite{Lilja2023DataLeakage} \cite{TFDV_Documentation} .\\

In the following, we provided Tables, \ref{tab:dataset_overview_wide_nolinks},\ref{tab:dataset_assessment_wide},  and \ref{tab:vv_methods_kpis},  to show a high-level comparison between latest V\&V approaches in dataset. 
Table \ref{tab:dataset_overview_wide_nolinks} summarizes representative autonomous driving dataset and dataset assurance literature by characterizing each work in terms of the dataset modality/type, the ODD and scenario coverage it targets, and the dataset safety properties addressed from an ISO/PAS 8800 perspective as mentioned in the section ~\ref{sec:Dataset safety properties}.
This synthesis is intended to provide a structured reference for comparing how different dataset families (real world multimodal benchmarks, simulation datasets, cooperative perception datasets, and specialized studies such as anomaly/OOD and leakage analyses) contribute to evidence relevant for safety assured learning-enabled systems.

The table \ref{tab:dataset_overview_wide_nolinks} columns provide a structured view of datasets and related studies from a safety assurance perspective.\\
The \textit{Paper/Dataset} column identifies the primary scholarly reference and, where applicable, the dataset itself. It distinguishes between dataset releases, simulation platforms, and analytical studies, clarifying whether safety evidence is produced by dataset creators or by external evaluations.\\
The \textit{Type of Data} Used column classifies the data modality and provenance, such as real‑world multimodal sensing, synthetic simulation, cooperative V2X data, or anomaly‑focused collections. This distinction reflects differences in achievable data quality assurance and verification strength.\\
The \textit{ODD / Scenario Coverage} column summarizes the operational contexts represented in the dataset, including environment, time, weather, geography, and interaction complexity. This links dataset composition to intended use and highlights potential gaps in safety‑critical scenarios.\\
The \textit{Safety Properties Addressed} column indicates which ISO/PAS 8800 dataset‑related safety properties are explicitly supported by documented evidence. It does not claim compliance but records whether sufficient information is available to support a standards‑aligned safety assurance argument.

Table \ref{tab:dataset_overview_wide_nolinks} reveals a consistent pattern across the surveyed literature: widely used real world benchmark datasets (e.g., KITTI, nuScenes, Waymo) most strongly and explicitly document dataset assurance elements that support reproducible benchmarking and auditability, particularly integrity, correctness/fidelity, traceability, and verifiability, through calibrated/synchronized data collection, structured dataset packaging, and publicly released devkits and evaluation protocols.
 \\

\textbf{Evidence-Based Property Assignment Methodology:}
To ensure consistency and prevent over-claiming of dataset safety properties, we adopted an evidence-trigger rubric based on the property definitions provided in ISO/PAS 8800. A given property was considered addressed only if the corresponding paper, dataset, or tool presented at least one explicit mechanism or a versionable artifact that aligned with the property’s requirements.
Table \ref{tab:dataset_assessment_wide} provides the Evidence-trigger mapping of ISO/PAS 8800 dataset safety properties per reference.\\

\subsubsection{Abbreviations used in the Table \ref{tab:dataset_assessment_wide}}

\begin{itemize}
    \item Explicit (E) = The paper, dataset, or tool explicitly documents evidence that aligns with the ISO/PAS~8800 definition of the safety property.\\
\item Implicit/Partial (P) = Some supporting evidence is present, but the property is not fully specified, quantified, or auditable in the publication.\\
\item No evidence (—) = The safety property is not evidenced in the cited reference.\\
\item Not applicable (N/A) = The safety property is not applicable to the type of reference (e.g., tool, simulator, or survey).
\end{itemize}

%%%%%%%%%%%%%%
\begin{table}[h!]
\caption{Challenges in Dataset Verification and Validation}
\scriptsize
\centering
\begin{tabular}{|p{1.5cm}|p{5cm}|}
\hline
\textbf{Challenge} & \textbf{Description} \\
\hline
Lack of Precise AI Safety Requirements & AI systems need to identify and quantify high-level semantic concepts (e.g., road objects, traffic signals) which lack precise definitions and are difficult to mathematically describe. \\
\hline
Versatile Inputs & AI systems receive inputs from diverse sources (e.g., radar, LiDAR, camera) requiring high-dimensional representations with complex constraints, making traditional input coverage methods incomplete or expensive. \\
\hline
Complex Architectures & AI systems, especially those using DNNs, have complex architectures with millions of parameters tuned during training, leading to scalability issues in verification for optimal performance and safety requirements. \\
\hline
Heuristic-Based Training & Training involves heuristics to optimize cost functions, which may result in locally optimal parameters without achieving global optimization, causing AI errors. Verification checks if parameters meet safety requirements. \\
\hline
Large Dataset Dependency & AI systems rely on large datasets for reliable performance. Validating the completeness and accuracy of these datasets is challenging due to the complexity of the input space. \\
\hline
Erroneous Behaviour & Data-driven AI systems can exhibit unpredictable errors due to spurious correlations, limiting performance prediction based on training data or model review. Lack of explainability exacerbates this issue. \\
\hline
Structural Coverage Limitations & Due to unspecified details and parameter dependencies, both black box and white box coverage metrics are limited in evaluating performance across the entire input space. \\
\hline
Environmental Stability & Small changes in the input space or function can lead to AI errors. Re-training can unpredictably impact previously verified properties. \\
\hline
Local Optimum Training & Training with insufficient examples can lead to local optima, resulting in behaviours misaligned with desired outcomes. \\
\hline
\end{tabular}
\label{table:challenges}
\vspace{0.5cm}
\end{table}
%%%%%%%%%%%%%%%%%%

\subsubsection{Explanation of Safety Properties with respect to each referenced work}
In the following, we examine each paper referenced in Table~\ref{tab:dataset_assessment_wide} and justify why a given safety property is classified as explicit or implicit for that reference.\\
\begin{itemize}
    \item  \textit{KITTI} \cite{Geiger2012KITTI}\
KITTI provides real‑world multimodal driving data with calibrated sensors and benchmark ground truth, supporting correctness, integrity, traceability, and verifiability through standardized tasks and evaluation infrastructure. Accuracy, independence, completeness, and representativeness are only partially supported, as the cited description does not report quantified label audits, leakage controls, or explicit coverage targets. Temporality is not addressed.\

\item \textit{nuScenes }\cite{caesar2020nuscenes}\
nuScenes offers synchronized multimodal sensor data with structured metadata, benchmarks, and tooling, supporting correctness, integrity, traceability, and verifiability. Accuracy, completeness, representativeness, and temporality are partial due to missing quantified QA metrics, coverage targets, and explicit drift policies. Independence is not evidenced.\

\item \textit{Waymo Open Dataset} \cite{Sun2020Waymo}\
Waymo provides large‑scale calibrated real‑world data with evaluation code and baselines, supporting correctness, integrity, traceability, and verifiability. Accuracy, completeness, representativeness, and temporality remain partial, as explicit audit rates, coverage targets, and drift strategies are not documented. Independence is not addressed.\

\item \textit{CARLA Simulator} \cite{dosovitskiy2017carla}\
CARLA supports traceability and verifiability through reproducible scenario configuration and privileged ground truth. Completeness, representativeness, integrity, and temporality are user‑configurable and therefore partial. Accuracy and independence are not applicable to simulator‑generated ground truth.\

\item \textit{LGSVL / SVL Simulator }\cite{rong2020lgsvl}\
LGSVL enables repeatable scenario execution and controlled testing, supporting traceability and verifiability. Other safety properties are user‑dependent and treated as partial capabilities. Accuracy and independence are not applicable to a fixed dataset.\

\item \textit{DAIR V2X }\cite{Yu2022DAIRV2X,DAIRV2X_Website}\
provides cooperative real‑world data with calibration and synchronization artifacts, supporting correctness, integrity, traceability, and verifiability. Accuracy, completeness, representativeness, and temporality are partial due to missing quantified audits and coverage targets. Independence is not evidenced.\

\item \textit{Anomaly Dataset Survey }\cite{Bogdoll2023AnomalySurvey}\
The survey explicitly addresses completeness for anomaly and OOD conditions and supports traceability and verifiability at the catalog level. Representativeness and temporality vary across datasets and are partial. Other properties are not applicable.\

\item \textit{Leakage Study} \cite{Lilja2023DataLeakage}\
The study directly evidences independence by identifying and mitigating geographic leakage, and supports traceability and verifiability through reproducible split definitions and comparative evaluation. Other properties are not applicable.\

\item \textit{TensorFlow Data Validation }\cite{TFDV_Documentation}\
TFDV supports integrity, temporality, traceability, and verifiability through schema‑based validation, drift detection, and versioned statistics. Other properties are not applicable, as TFDV is a validation tool rather than a dataset.

\end{itemize}
%%%%%%%%%%%%%%%%%%

\begin{table*}[t]
\centering
\caption{Verification and Validation Practices Across Datasets, Simulators, and Data Validation Tooling}
\label{tab:vv_methods_kpis}
\scriptsize
\setlength{\tabcolsep}{5pt}
\renewcommand{\arraystretch}{1.15}
\begin{tabular}{p{3.6cm} p{4cm} p{4cm} p{4cm}}
\hline
\textbf{Paper / Dataset} &
\textbf{Verification Methods} &
\textbf{Validation Methods} &
\textbf{KPIs} \\
\hline

KITTI \cite{Geiger2012KITTI} &
Sensor calibration and synchronization evidence; benchmark ground-truth construction checks; dataset packaging and split-definition checks. &
Validation for intended use as a perception benchmark via standardized evaluation on held-out splits (e.g., detection, tracking, odometry, optical flow within the dataset ODD). &
Task-specific benchmark metrics computed by the official evaluation protocol; acceptance typically baseline-, leaderboard-, or project-target-relative. \\

nuScenes \cite{caesar2020nuscenes} &
Annotation schema consistency checks; dataset statistics checks; devkit-based evaluation reproducibility checks; calibration and metadata consistency. &
Validation for intended use as a multimodal perception benchmark via standardized detection and tracking evaluation on defined splits (dataset ODD). &
Official nuScenes detection and tracking metrics computed via the devkit; acceptance typically baseline-relative or program-defined targets. \\

Waymo Open Dataset \cite{Sun2020Waymo} &
Sensor synchronization and calibration evidence; label format and specification checks; official evaluation code conformance checks. &
Validation for intended use as a perception benchmark via standardized detection and tracking evaluation and cross-region generalization assessment on held-out splits. &
Detection and tracking metrics computed via the official evaluation pipeline; acceptance typically baseline- or target-relative, with generalization assessed across subsets. \\

CARLA Simulator \cite{dosovitskiy2017carla} &
Verification-by-construction for simulator-generated ground truth; verification of scenario and sensor configurations against specified test requirements. &
Validation for intended use in simulation by executing defined scenarios and assessing system behavior against scenario-level acceptance criteria (simulation ODD). &
Scenario outcome KPIs (e.g., collisions, infractions, route completion); thresholds defined per scenario or test specification (pass/fail). \\

LGSVL / SVL Simulator \cite{rong2020lgsvl} &
Verification of configured sensor models and parameters and scenario setup against specified test requirements; repeatability checks. &
Validation for intended use in simulation via repeatable scenario execution with integrated AV stacks and assessment against scenario-level pass/fail criteria. &
Scenario-specific KPIs (e.g., collisions, rule violations, completion); thresholds defined by test specifications and stack-level objectives. \\

DAIR-V2X \cite{Yu2022DAIRV2X,DAIRV2X_Website} &
Temporal and spatial synchronization and calibration checks; timestamp consistency checks; cooperative dataset structuring and labeling consistency checks. &
Validation for intended use as a cooperative perception benchmark via standardized cooperative 3D detection evaluation under synchronous and asynchronous conditions. &
Cooperative detection metrics computed under the benchmark protocol; acceptance typically baseline-relative and condition-specific (e.g., by asynchrony setting). \\

Anomaly Dataset Survey \cite{Bogdoll2023AnomalySurvey} &
Survey-level characterization criteria to confirm dataset attributes (e.g., ground truth type, anomaly definition, context metadata, licensing) for consistent categorization. &
Validation of suitability for anomaly and out-of-distribution method evaluation by mapping datasets to tasks, ground-truth availability, and evaluation protocols. &
Dataset comparison attributes and anomaly-evaluation KPIs (dataset-dependent); thresholds typically risk-driven (e.g., false-alarm trade-offs) rather than fixed dataset-wide limits. \\

Leakage Study \cite{Lilja2023DataLeakage} &
Geographic overlap (leakage) analysis across splits; enforcement of disjoint-split constraints to satisfy independence requirements. &
Validation for intended use of fair generalization assessment by re-evaluating methods under leakage-free (geographically disjoint) splits. &
Overlap KPIs (e.g., distance-based overlap rates) and performance deltas under corrected splits; thresholds implemented as hard split constraints. \\

TFDV \cite{TFDV_Documentation} &
Schema-based example validation and anomaly detection confirming conformance to specified data expectations (types, ranges, presence). &
Validation support for intended use by checking training--serving skew and monitoring drift over time windows to ensure continued data suitability. &
Statistics-based anomaly, skew, and drift comparators; thresholds user-configured via schema and comparator settings. \\

\hline
\end{tabular}
\end{table*}
%%%%%%%%%%%%%%%%%%%%%%

At the same time, several other ISO/PAS 8800 dataset safety properties—including accuracy, completeness, representativeness, independence, and temporality—are frequently treated only implicitly in dataset releases, because many publications describe “high quality” or “diverse” data without providing audit grade label quality statistics, explicit scenario/ODD coverage targets, formal distribution/bias analyses against intended operational exposure, explicit leakage controls, or time based drift/refresh strategies. Simulation platforms (e.g., CARLA and LGSVL/SVL) exhibit a complementary profile: they are explicitly strong on traceability and verifiability due to reproducible scenario configurations and privileged ground truth, while evidence for other dataset safety properties (e.g., completeness, representativeness, temporality, integrity) depends on scenario design and sampling choices rather than being guaranteed by the platform itself. The table further indicates that explicit evidence for independence is most clearly provided by targeted leakage studies that quantify overlap and propose disjoint split strategies, reinforcing the broader survey conclusion that ISO/PAS 8800 aligned dataset assurance requires property specific, objective evidence artifacts and monitoring mechanisms rather than relying on benchmark performance claims alone.

\subsubsection{Review on Dataset V\&V Practices}
Table \ref{tab:vv_methods_kpis} consolidates dataset-level verification and validation practices reported across representative autonomous driving dataset releases, simulation platforms, methodological studies, and dataset assurance tooling.\\
The table illustrates that benchmark datasets (e.g., KITTI, nuScenes, Waymo) primarily implement verification through calibration and synchronization artifacts, as well as standardized evaluation tooling (development kits and evaluation code) that enforce format and protocol constraints. Validation, in these cases, is operationalized via benchmark-oriented task evaluation, employing standardized metrics and published baselines. Simulation platforms (e.g., CARLA, LGSVL/SVL) achieve verification by means of configuration (sensor/scenario specifications and repeatability), and validation through controlled scenario execution with scenario-level metrics, thereby supporting their intended use as simulation-based evaluation environments. In cooperative perception datasets (e.g., DAIR-V2X), there is an increased emphasis on verification regarding temporal alignment and asynchrony, while validation is demonstrated through cooperative benchmark tasks and baselines.
Specialized assurance studies and tools serve as complements to benchmark datasets. For example, the leakage study validates dataset split independence via objective geographic overlap analysis and construction of disjoint splits. TensorFlow Data Validation (TFDV) supports verification through schema and statistical anomaly checks, and provides validation support for intended use by monitoring training-serving skew and data drift over time windows to ensure ongoing suitability during operation.
Finally, Table \ref{tab:vv_methods_kpis} explicitly details that KPI thresholding practices are typically:\\
(i) baseline/leaderboard-relative for benchmarks,\\
(ii) scenario-specific pass/fail for simulation,\\
(iii) enforced as hard constraints for independence/leakage mitigation, or\\
(iv) user-configured comparators for drift and anomaly monitoring. This highlights how evidence is operationalized into auditable decision criteria, rather than being stated only qualitatively.

\section{Conclusion and Future Direction of Research}
In this paper, we have presented comprehensive guidelines for the development of end-to-end (E2E) AI datasets tailored specifically for autonomous driving applications. By closely aligning our approach with contemporary automotive safety standards, including ISO 21448 (SOTIF) and ISO/PAS 8800, we outlined explicit methods to systematically derive dataset requirements from AI safety requirements. Our framework emphasizes maintaining rigorous traceability between these safety requirements and dataset specifications, thereby reinforcing the reliability and compliance of autonomous driving systems.

 We detailed critical dataset safety properties outlined in ISO/PAS 8800, such as completeness and independence, and provided practical examples illustrating common data insufficiencies that can compromise safety. By incorporating systematic safety analyses at every stage of the dataset lifecycle—requirement definition, dataset design, and implementation—we have underscored the importance of continuous verification and validation processes. These processes are essential to ensure datasets comprehensively cover operational domains, avoid information leakage, and adhere to the necessary safety criteria.
 
 Despite these advancements, significant challenges persist in dataset development for autonomous driving. A primary challenge highlighted in this study is the agility required for dataset development. Autonomous driving AI systems typically evolve rapidly through agile development processes. Consequently, the dataset development and associated safety evaluations must match this rapid pace, which demands innovative methods to streamline safety assessments without compromising rigor or compliance. Meeting this challenge is essential for maintaining effective safety validation in rapidly evolving autonomous driving technologies.
 
 Looking forward, several promising directions for future research have emerged. One critical area involves developing methods for rapidly constructing AI datasets suitable for end-to-end AI architectures within Advanced Driver Assistance Systems (ADAS). The swift evolution of ADAS demands datasets that can be quickly and flexibly expanded or modified while maintaining strict adherence to safety and quality standards. Addressing this demand necessitates developing scalable and automated approaches for dataset collection, annotation, verification, and validation.
 
 Another vital future research area pertains to security considerations in dataset development for ADAS. As datasets form the backbone of AI systems, ensuring their security against adversarial attacks, unauthorized modifications, or privacy breaches is imperative. Future research should prioritize the development of robust security frameworks that safeguard datasets through stringent access control, secure data handling practices, and comprehensive threat modeling. Exploring advanced security measures, including encryption techniques and secure data transmission protocols tailored explicitly for automotive datasets, will further enhance the integrity and trustworthiness of autonomous driving systems.

%\clearpage

%\input{include/related_work}

% \input{include/LLP.tex}

%\input{}

% \input{include/architecture.tex}

% \input{include/results.tex}

\bibliographystyle{IEEEtran}
\bibliography{references/egbib}

\end{document}